\documentclass[11pt,letterpaper]{article}
\usepackage[T1]{fontenc}
\usepackage[utf8]{inputenc}
\usepackage{lmodern}

\usepackage[margin=1in]{geometry}
\usepackage{amsmath,amssymb}
\usepackage{graphicx,booktabs,array,multirow}
\usepackage{xcolor,helvet,etoolbox}
\usepackage[protrusion=false]{microtype}
\usepackage{caption}
\usepackage{hyperref}
\hypersetup{
  hidelinks,
  pdftitle={FARR: Forecast-Aware Residual Refinement with Matched Endpoint Evidence for Long-Horizon Time-Series Forecasting},
  pdfauthor={Yiyang Zhang, Yuli Zheng, Bin Li, Meimei Zhang, Qiang Sun},
  pdfsubject={Long-horizon time-series forecasting},
  pdfkeywords={long-horizon forecasting, endpoint evidence, residual refinement, horizon-wise gating, language-model-based forecasting},
  pdfcreator={LaTeX}
}
\DeclareCaptionFont{tabletext}{\fontsize{10.1}{12.5}\selectfont}
\patchcmd{\thebibliography}{\section*{\refname}}{\section{\refname}}{}{}
\makeatletter
\renewcommand{\@maketitle}{%
  \newpage\null\vskip 0.1em
  \begin{center}
    \let\footnote\thanks
    {\fontsize{16}{19}\selectfont\bfseries\MakeUppercase{\@title}\par}
    \vskip 0.7em
    {\large\lineskip 0.4em\begin{tabular}[t]{c}\@author\end{tabular}\par}
  \end{center}
  \par\vskip 0.25em
}
\makeatother
\title{FARR: Forecast-Aware Residual Refinement with Matched Endpoint Evidence for Long-Horizon Time-Series Forecasting}
\author{%
  Yiyang Zhang\textsuperscript{1}, Yuli Zheng\textsuperscript{2}, Bin Li\textsuperscript{3},
  Meimei Zhang\textsuperscript{4}, Qiang Sun\textsuperscript{1,}\thanks{\fontsize{10.1}{12.5}\selectfont Corresponding author.}\\[0.7em]
  \begin{tabular}{c}
  \fontsize{10.1}{12.5}\selectfont\textsuperscript{1}Nantong University, Nantong, China\\
  \fontsize{10.1}{12.5}\selectfont\textsuperscript{2}Faculty of Mathematics and Statistics, University of Ottawa, Ottawa, Canada\\
  \fontsize{10.1}{12.5}\selectfont\textsuperscript{3}Shenzhen Institute of Advanced Technology, Chinese Academy of Sciences, Shenzhen, China\\
  \fontsize{10.1}{12.5}\selectfont\textsuperscript{4}Amazon.com Inc., Seattle, WA, USA
  \end{tabular}}
\date{}

\begin{document}
\maketitle
\begin{center}
\begin{minipage}{0.95\textwidth}
\fontsize{10.1}{12.5}\selectfont\centering
This work has been submitted to IEEE for possible publication. Copyright may be transferred without notice, after which this version may no longer be accessible.
\end{minipage}
\end{center}

\begin{abstract}
Large language models use auxiliary conditioning to incorporate additional information into time-series forecasts. However, auxiliary conditioning does not consistently improve accuracy and can increase forecasting error when its influence is poorly matched to the observed temporal context. We propose Forecast-Aware Residual Refinement (FARR), which learns to correct forecasts using endpoint evidence. FARR obtains a valid forecast under the standard condition and a null forecast under a predefined null condition. Their difference reveals how conditioning changes the prediction. A shared residual refiner combines this endpoint evidence with historical features to estimate the remaining error of the valid forecast. Horizon-wise gating then controls the contribution of the predicted residual to the final output. Experiments on four long-horizon forecasting benchmarks show that FARR achieves the best dataset-average MSE and MAE among the compared methods. Code will be released at \url{https://github.com/blossomrichzyy-star/farr-forecast-aware-residual-refinement}.
\end{abstract}
\noindent\textbf{Index Terms---} Long-horizon forecasting, endpoint evidence, residual refinement, horizon-wise gating, language-model-based forecasting

\section{Introduction}
Language-model-based forecasting incorporates auxiliary information into time-series prediction through conditioning. Time-LLM adapts numerical inputs to pretrained language models through reprogramming~\cite{timellm}. $\mathrm{S}^{2}$IP-LLM learns prompts informed by the language embedding space~\cite{s2ip}, while Deep TPC introduces temporal information at multiple decoder layers~\cite{deeptpc}. AutoTimes further adapts autoregressive language modeling to time-series forecasting~\cite{autotimes}. These methods differ in how they construct conditioning, whose forecasting benefit depends on the data.

In parallel, forecasting architectures have improved the representation of temporal structure. DLinear demonstrates the strength of simple decomposition-based prediction~\cite{dlinear}. Informer reduces the cost of long-sequence attention~\cite{informer}, and Autoformer incorporates decomposition and auto-correlation~\cite{autoformer}. FEDformer introduces frequency-domain operations~\cite{fedformer}, whereas PatchTST learns from temporal patches~\cite{patchtst}. iTransformer models dependencies between variables~\cite{itransformer}, and TimeMixer combines information across temporal scales~\cite{timemixer}. Periodic modeling in CycleNet~\cite{cyclenet}, efficient attention in Ister~\cite{ister}, and spectral filtering before attention~\cite{filterattend} provide further improvements. Large-scale pretraining offers another direction, as illustrated by Chronos~\cite{chronos}, TimesFM~\cite{timesfm}, and Time-MoE~\cite{timemoe}; DistilTS transfers forecasting capabilities to smaller models~\cite{distilts}.

For language-model-based forecasters, a related line of work improves the interaction between numerical observations and auxiliary representations. CALF uses cross-modal fine-tuning~\cite{calf}, and TimeCMA aligns time-series and language representations~\cite{timecma}. Retrieval augmentation instead supplies relevant historical examples~\cite{raft}. Nevertheless, stronger conditioning does not imply that every resulting forecast is more accurate. Controlled studies also show that the benefits of language-model components depend on the forecasting setting~\cite{usefulllm}.

\begin{figure}[htbp]
\centering
\includegraphics[width=0.95\textwidth]{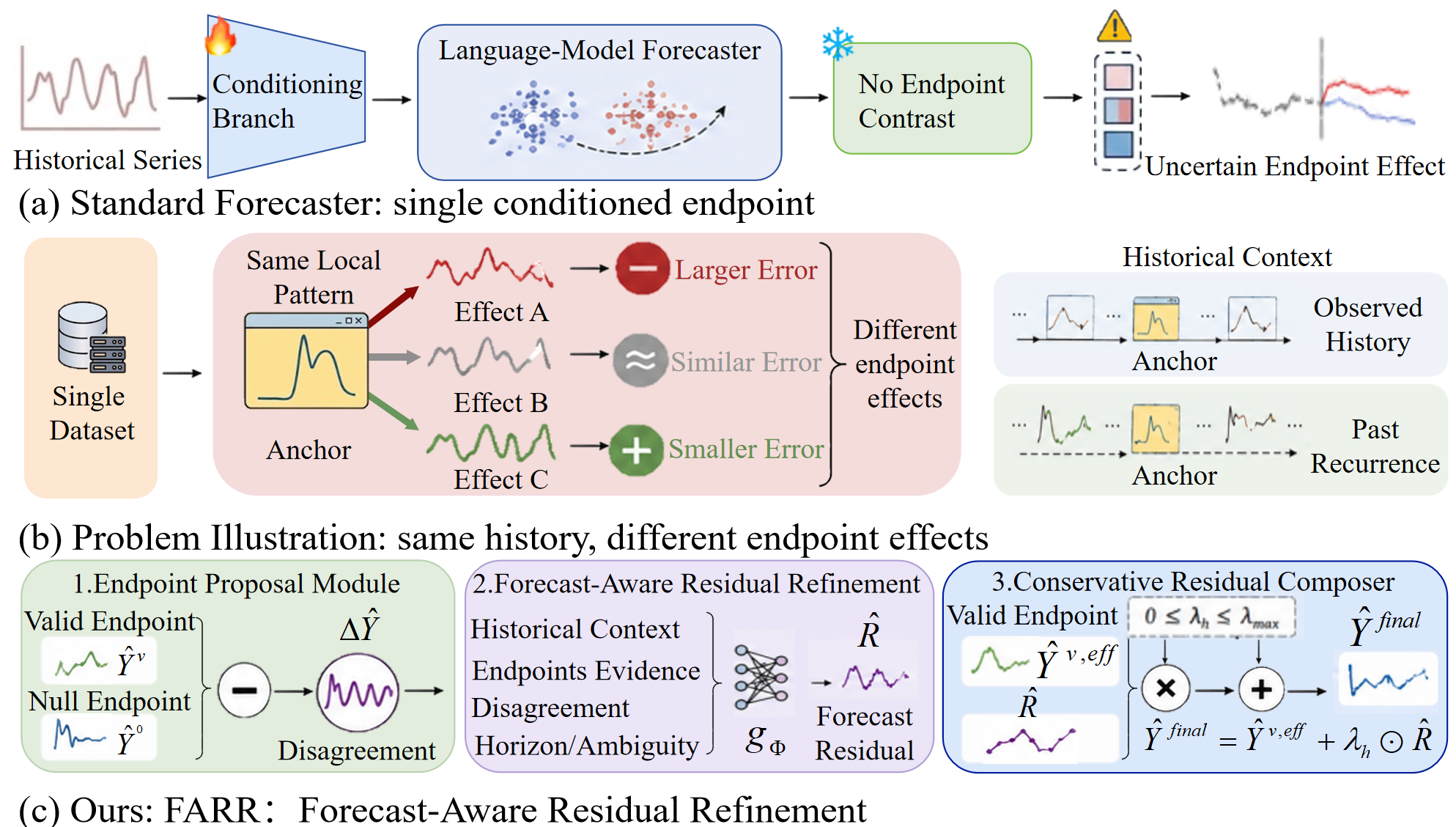}
\caption{Motivation of FARR. (a) A single conditioned forecast does not reveal the prediction change relative to a null condition. (b) Conditioning can have different effects on forecasting error. (c) FARR combines endpoint evidence with historical features for residual refinement and applies horizon-wise gating before adding the residual to the effective valid forecast.}
\label{fig:motivation}
\end{figure}

Figure~\ref{fig:motivation} illustrates this distinction. A standard pipeline produces a single conditioned forecast, leaving its change relative to a null condition unobserved. Conditioning can increase, barely change, or decrease prediction error. The difference between two forecasts is observable at inference, but its relationship to future error must be learned from training data. Historical context is therefore needed to determine how this difference should influence the final prediction.

We address this problem through residual refinement. RefineBridge demonstrates that forecast outputs can serve as starting points for a learned refinement process~\cite{refinebridge}. FARR focuses on the additional information provided by two forecasts generated under different conditions. It combines their difference with historical features to predict a residual relative to the valid forecast. A horizon-wise gate controls how strongly this residual changes the output.

\begin{samepage}
Our contributions are threefold:
\begin{itemize}
\item We introduce endpoint evidence from valid and null forecasts, making the prediction change associated with conditioning available to residual refinement.
\item We develop a shared residual refiner and horizon-wise gating that use endpoint evidence and historical features to correct the valid forecast.
\item We evaluate FARR on four forecasting benchmarks, obtaining the best dataset-average MSE and MAE among the compared methods. Ablations examine endpoint inputs, historical features, residual refinement, and forecast adjustment.
\end{itemize}
\end{samepage}

\section{Method}
\subsection{Problem formulation}
\begin{figure}[htbp]
\centering
\includegraphics[width=\textwidth]{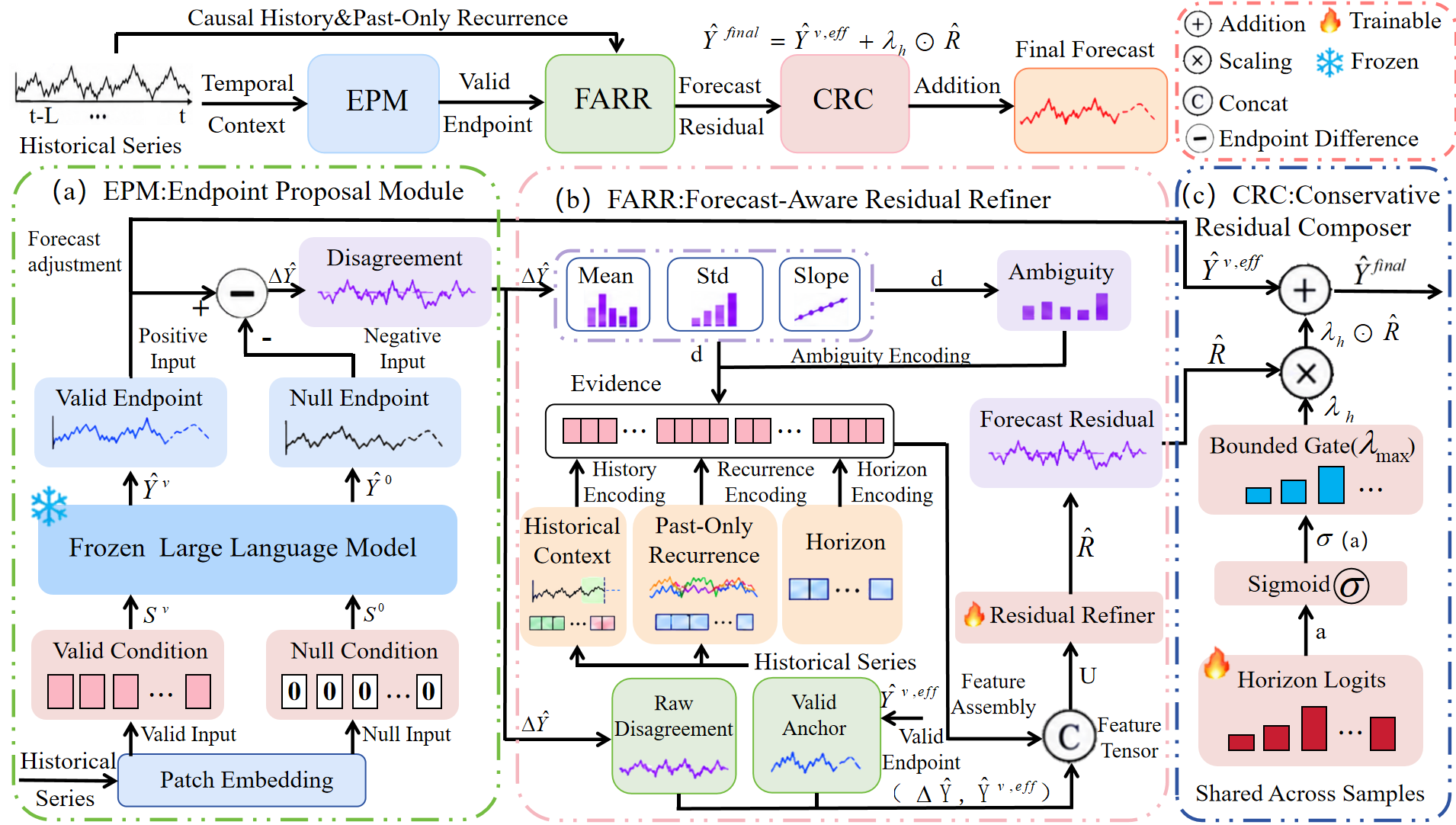}
\caption{FARR architecture. EPM obtains matched endpoints, and forecast adjustment yields the effective valid forecast. The residual refiner combines endpoint evidence with historical context in $U$. CRC applies horizon-wise gating and adds the predicted residual to $\widehat{Y}^{v,\mathrm{eff}}$.}
\label{fig:framework}
\end{figure}

Let $X_i\in\mathbb R^{L\times C}$ denote the historical observations at forecast origin $i$, where $L$ is the input length and $C$ is the number of channels. The target $Y_i\in\mathbb R^{H\times C}$ contains the next $H$ observations. A frozen forecaster $F_\theta$ receives the history under a specified conditioning operation. FARR consists of an Endpoint Proposal Module (EPM), a residual refiner, and a Conservative Residual Composer (CRC), as shown in Figure~\ref{fig:framework}. Refinement uses observed history and matched endpoint forecasts, providing a common refinement interface for forecasters with a compatible conditioning pathway.

\subsection{Endpoint evidence}
For a shared history, EPM evaluates the frozen forecaster under standard conditioning $S_i^v$ and a fixed reference (null) operation $S_i^0$:
\begin{equation}
\widehat{Y}_i^v=F_\theta(X_i,S_i^v),\qquad
\widehat{Y}_i^0=F_\theta(X_i,S_i^0).
\label{eq:endpoints}
\end{equation}
Both passes share the checkpoint, history, preprocessing, output settings, and sample order; only conditioning changes. Here, ``valid'' denotes standard conditioning, not the validation split. We define
\begin{equation}
\Delta\widehat{Y}_i=\widehat{Y}_i^v-\widehat{Y}_i^0.
\label{eq:delta}
\end{equation}
The forecasts and their difference constitute \emph{endpoint evidence}; $\Delta\widehat{Y}_i$ is the \emph{endpoint disagreement} induced by the reference operation.

The reference operation acts on a designated auxiliary conditioning pathway. The zero symbols in Figure~\ref{fig:framework} denote this intervention, with historical inputs unchanged. The refiner uses only the resulting forecasts and observed history.

\subsection{Residual refinement and forecast adjustment}
Residual architectures model successive corrections in time-series forecasting~\cite{nbeats}. The refiner receives a fixed 18-dimensional feature vector $u_{i,h,c}$. Indices 0--7 describe history: the latest value, mean, scale, slope, first-difference scale, local volatility, motif similarity, and ambiguity. Indices 8--9 are the null and valid forecasts. Indices 10--16 describe their disagreement: signed difference, absolute difference, history-scale-normalized difference, first-order difference, smoothness, relative displacement, and sign consistency. Index 17 is the normalized horizon coordinate. These 18 features are deterministic functions of history, forecasts, and horizon position; the encodings in Figure~\ref{fig:framework} denote these fixed computations. Their mapping to a residual is learned.

History statistics use the latest 168 observations. The history scale is the population standard deviation, floored at $10^{-4}$; slope is $(x_L-x_{L-167})/167$. The first-difference scale uses the same normalization and floor. Local volatility is the mean absolute first difference. Motif similarity is the cosine similarity between the latest and preceding 24-point windows, representing past-only recurrence in Figure~\ref{fig:framework}. Ambiguity is $\mathrm{clip}(v/\sigma+1-m,0,10)$, with local volatility $v$, history scale $\sigma$, and motif similarity $m$. For endpoint disagreement $\Delta_h$, the derived quantities are
\begin{equation}
d_h^{(1)}=\Delta_h-\Delta_{h-1},\qquad
q_h=\mathrm{clip}\!\left(\frac{|d_h^{(1)}|}{|\Delta_h|+10^{-3}},0,10\right),
\label{eq:derived1}
\end{equation}
\begin{equation}
r_h=\frac{|\Delta_h|}{|\widehat Y_h^0|+10^{-3}},\qquad
k=\left|\frac{1}{H}\sum_{h=1}^{H}\mathrm{sign}(\Delta_h)\right|,
\label{eq:derived2}
\end{equation}
where $d_1^{(1)}=0$. The horizon coordinate is linearly spaced from 0 to 1. Concatenating these features gives $U\in\mathbb R^{B\times H\times18}$, with $B$ single-channel windows. One MLP with two hidden layers is shared across prediction steps and channels:
\begin{equation}
\widehat R_{i,h,c}=\tanh\!\left(g_\Phi(u_{i,h,c})\right),\qquad
g_\Phi:18\rightarrow128\rightarrow128\rightarrow1.
\label{eq:refiner}
\end{equation}
Both hidden layers use GELU; the output layer is zero-initialized. Thus $\widehat R\in[-1,1]$ in standardized space.

Stage 2 applies a forecast adjustment to the valid forecast relative to the null forecast:
\begin{equation}
\widehat Y_{i,h,c}^{v,\mathrm{adj}}=\widehat Y_{i,h,c}^{0}
+[1+0.10\tanh(s_h)]\Delta\widehat Y_{i,h,c}
+0.10\tanh(b_h).
\label{eq:adjustment}
\end{equation}
The horizon-wise vectors $s,b\in\mathbb R^H$ are zero-initialized and shared across samples and channels. Stage 2 recomputes endpoint-dependent features with $\widehat{Y}^{v,\mathrm{adj}}$ and disagreement $\widehat{Y}^{v,\mathrm{adj}}-\widehat{Y}^{0}$, using residual target $Y-\widehat{Y}^{v,\mathrm{adj}}$. Gradients pass through the adjusted forecast. In Stage 1, the reference is the original valid forecast. We denote this stage-specific reference by the \emph{effective valid forecast} $\widehat{Y}^{v,\mathrm{eff}}$.

\subsection{Horizon-wise gating}
CRC controls the predicted residual through a learnable coefficient for each prediction step:
\begin{equation}
\lambda_h=\lambda_{\max}\,\mathrm{sigmoid}(a_h),\qquad
\lambda_{\max}=0.20,\quad h=1,\ldots,H.
\label{eq:gate}
\end{equation}
The logits are initialized at $a_h=-6$ and shared across samples and channels. The final forecast is
\begin{equation}
\widehat Y^{\mathrm{final}}_{i,h,c}=\widehat Y^{v,\mathrm{eff}}_{i,h,c}+\lambda_h\widehat R_{i,h,c}.
\label{eq:final}
\end{equation}
The gated residual is bounded by $[-0.20,0.20]$ relative to the effective valid forecast in standardized space. Sample-dependent corrections come from the residual refiner. The refiner has 19,073 parameters; gating and forecast adjustment add $3H$, giving 19.4K--21.2K trainable parameters in total.

\subsection{Training objective}
\label{sec:selection}
FARR is trained with forecast and residual losses and two residual regularizers:
\begin{equation}
\begin{split}
\mathcal L={}&\mathrm{SmoothL1}(\widehat{Y}^{\mathrm{final}},Y)
+\alpha_r\,\mathrm{SmoothL1}(\widehat{R},Y-\widehat{Y}^{v,\mathrm{eff}})\\
&+\beta_s\mathcal L_{\mathrm{smooth}}+\beta_{\ell2}\mathcal L_{\ell2}.
\end{split}
\label{eq:loss}
\end{equation}
We use SmoothL1 (Huber loss~\cite{huber} with unit transition), $\alpha_r=0.10$, and $\beta_s=\beta_{\ell2}=0.01$. For $B$ single-channel windows $j=(i,c)$,
\begin{equation}
\mathcal L_{\mathrm{smooth}}=\frac{\sum_{j,h}(\widehat R_{j,h+1}-\widehat R_{j,h})^2}{B(H-1)},\qquad
\mathcal L_{\ell2}=\frac{\sum_{j,h}\widehat R_{j,h}^{2}}{BH}.
\label{eq:regularizers}
\end{equation}
The smoothness sum uses $h<H$. Both regularizers act before gating. Stage 1 trains the refiner and gate. Stage 2 starts only if Stage 1 improves both validation MSE and MAE over the valid forecast, and jointly trains the forecast adjustment. Each checkpoint update requires both metrics to strictly improve. Stage-2 candidates must also improve both metrics over the valid forecast; otherwise Stage 1 is retained, with the valid forecast as fallback. All selection uses validation data; the frozen candidate is evaluated on the test split using Eq.~\eqref{eq:final}.

\section{Experiments}
\noindent\textbf{Experimental setup.} We evaluate FARR on ETTh1, Weather, Traffic, and Electricity (ECL), using input length $L=672$ and prediction horizons $H\in\{96,192,336,720\}$. A StandardScaler is fitted on the training split and applied unchanged to validation and test data. All locally evaluated MSE and MAE, including the local valid anchor and FARR, are computed in standardized space. Dataset averages weight the four horizons equally. Endpoint forecasts use a frozen local checkpoint based on Qwen2.5-0.5B-Instruct, with label length 576 and token length 96. Table~\ref{tab:main} lists the forecasting baselines.

\begin{table}[!htbp]
\centering
\caption{Long-horizon forecasting results (MSE/MAE). \textcolor{red}{Red}/\textcolor{blue}{blue} mark the lowest/second-lowest values. Avg. averages the four displayed horizons; \textendash\ denotes unreported results. $*$ denotes our reproduced results.}
\label{tab:main}
\fontfamily{phv}\fontseries{mc}\fontsize{10.1}{12.5}\selectfont
\setlength{\tabcolsep}{0.85pt}
\renewcommand{\arraystretch}{1.16}
\begin{tabular}{@{}lc*{9}{!{{\color{black!25}\vrule width 0.2pt}}c}@{}}
\toprule
Dataset & $H$ & \shortstack{FARR\\(Ours)} & \shortstack{SE-LLM\\\cite{sellm}} & \shortstack{TimeMixer++\\\cite{timemixerpp}} & \shortstack{Time-MoE\\\cite{timemoe}} & \shortstack{TQNet\\\cite{tqnet}} & \shortstack{TimeCMA\\\cite{timecma}} & \shortstack{LLM-TS\\\cite{llmts}} & \shortstack{AutoTimes\\\cite{autotimes}} & \shortstack{Time-LLM\\\cite{timellm}}\\
\midrule
ETTh1 & 96 & 0.353/0.393 & \textcolor{blue}{0.352}/0.393 & 0.361/0.403 & \textcolor{red}{0.349}/\textcolor{red}{0.379} & 0.371/0.393 & 0.373/\textcolor{blue}{0.391} & 0.403/0.420 & 0.360/0.397 & 0.362/0.392 \\
 & 192 & \textcolor{blue}{0.379}/\textcolor{blue}{0.413} & \textcolor{red}{0.378}/\textcolor{red}{0.411} & 0.416/0.441 & 0.395/\textcolor{blue}{0.413} & 0.428/0.426 & 0.427/0.421 & 0.440/0.441 & 0.391/0.419 & 0.398/0.418 \\
 & 336 & \textcolor{red}{0.385}/\textcolor{red}{0.413} & \textcolor{blue}{0.391}/\textcolor{blue}{0.420} & 0.430/0.434 & 0.447/0.453 & 0.476/0.446 & 0.458/0.448 & 0.471/0.457 & 0.408/0.432 & 0.430/0.427 \\
 & 720 & \textcolor{red}{0.392}/\textcolor{red}{0.429} & \textcolor{blue}{0.402}/\textcolor{blue}{0.437} & 0.467/0.451 & 0.457/0.462 & 0.487/0.470 & 0.449/0.460 & 0.503/0.487 & 0.429/0.452 & 0.442/0.457 \\
 & Avg. & \textcolor{red}{0.377}/\textcolor{red}{0.412} & \textcolor{blue}{0.381}/\textcolor{blue}{0.415} & 0.419/0.432 & 0.412/0.427 & 0.441/0.434 & 0.427/0.430 & 0.454/0.451 & 0.397/0.425 & 0.408/0.424 \\
\midrule
Weather & 96 & \textcolor{red}{0.141}/\textcolor{red}{0.190} & 0.150/\textcolor{blue}{0.200} & 0.155/0.205 & 0.157/0.211 & 0.157/\textcolor{blue}{0.200} & 0.167/0.211 & 0.166/0.217 & 0.158/0.208 & \textcolor{blue}{0.147}/0.201 \\
 & 192 & \textcolor{red}{0.185}/\textcolor{red}{0.234} & 0.194/\textcolor{blue}{0.243} & 0.201/0.245 & 0.208/0.256 & 0.206/0.245 & 0.212/0.253 & 0.229/0.269 & 0.207/0.254 & \textcolor{blue}{0.189}/\textcolor{red}{0.234} \\
 & 336 & \textcolor{red}{0.242}/\textcolor{red}{0.277} & \textcolor{blue}{0.247}/0.285 & 0.248/0.285\textsuperscript{$*$} & 0.255/0.290 & 0.262/0.287 & 0.270/0.292 & 0.278/0.302 & 0.262/0.298 & 0.262/\textcolor{blue}{0.279} \\
 & 720 & \textcolor{red}{0.315}/\textcolor{red}{0.330} & 0.323/0.339 & \textcolor{blue}{0.316}/\textcolor{blue}{0.332}\textsuperscript{$*$} & 0.405/0.397 & 0.344/0.342 & 0.350/0.348 & 0.354/0.351 & 0.342/0.353 & 0.318/0.338\textsuperscript{$*$} \\
 & Avg. & \textcolor{red}{0.221}/\textcolor{red}{0.258} & \textcolor{blue}{0.229}/0.267 & 0.230/0.267\textsuperscript{$*$} & 0.256/0.289 & 0.242/0.269 & 0.250/0.276 & 0.257/0.285 & 0.242/0.278 & \textcolor{blue}{0.229}/\textcolor{blue}{0.263}\textsuperscript{$*$} \\
\midrule
Traffic & 96 & \textcolor{red}{0.344}/\textcolor{red}{0.240} & \textcolor{blue}{0.350}/\textcolor{blue}{0.243} & 0.392/0.253 & \textendash & 0.413/0.261 & \textendash & 0.587/0.315 & 0.369/0.257 & 0.362/0.248 \\
 & 192 & \textcolor{red}{0.367}/\textcolor{red}{0.252} & \textcolor{blue}{0.373}/\textcolor{blue}{0.254} & 0.402/0.258 & \textendash & 0.432/0.271 & \textendash & 0.612/0.326 & 0.394/0.268 & 0.390/0.288\textsuperscript{$*$} \\
 & 336 & \textcolor{red}{0.385}/\textcolor{red}{0.260} & \textcolor{blue}{0.390}/\textcolor{blue}{0.263} & 0.428/\textcolor{blue}{0.263} & \textendash & 0.450/0.277 & \textendash & 0.634/0.338 & 0.413/0.278 & 0.407/0.299\textsuperscript{$*$} \\
 & 720 & \textcolor{red}{0.420}/\textcolor{red}{0.279} & \textcolor{blue}{0.429}/0.285 & 0.441/\textcolor{blue}{0.282} & \textendash & 0.486/0.295 & \textendash & 0.640/0.351 & 0.449/0.299 & 0.430/0.288 \\
 & Avg. & \textcolor{red}{0.379}/\textcolor{red}{0.258} & \textcolor{blue}{0.386}/\textcolor{blue}{0.261} & 0.416/0.264 & \textendash & 0.445/0.276 & \textendash & 0.618/0.333 & 0.406/0.276 & 0.397/0.281\textsuperscript{$*$} \\
\midrule
ECL & 96 & \textcolor{red}{0.124}/\textcolor{red}{0.218} & \textcolor{blue}{0.130}/0.225 & 0.135/\textcolor{blue}{0.222} & \textendash & 0.134/0.229 & 0.143/0.238 & 0.167/0.271 & 0.140/0.236 & 0.131/0.224 \\
 & 192 & \textcolor{red}{0.136}/\textcolor{red}{0.233} & 0.148/0.242 & \textcolor{blue}{0.147}/\textcolor{blue}{0.235} & \textendash & 0.154/0.247 & 0.161/0.259 & 0.178/0.280 & 0.159/0.253 & 0.152/0.241 \\
 & 336 & \textcolor{red}{0.155}/0.251 & 0.164/0.259 & 0.164/\textcolor{red}{0.245} & \textendash & 0.169/0.264 & 0.169/0.261 & 0.198/0.302 & 0.177/0.270 & \textcolor{blue}{0.160}/\textcolor{blue}{0.248} \\
 & 720 & \textcolor{blue}{0.193}/\textcolor{red}{0.284} & 0.203/\textcolor{blue}{0.293} & 0.212/0.310 & \textendash & 0.201/0.294 & 0.219/0.315 & 0.233/0.344 & 0.216/0.303 & \textcolor{red}{0.192}/0.298 \\
 & Avg. & \textcolor{red}{0.152}/\textcolor{red}{0.247} & 0.161/0.255 & 0.165/\textcolor{blue}{0.253} & \textendash & 0.165/0.259 & 0.173/0.268 & 0.194/0.299 & 0.173/0.266 & \textcolor{blue}{0.159}/\textcolor{blue}{0.253} \\
\bottomrule
\end{tabular}
\end{table}

\subsection{Long-horizon forecasting results}
Table~\ref{tab:main} compares FARR with the forecasting baselines. Its dataset-average MSE/MAE is 0.3773/0.4120 on ETTh1, 0.2208/0.2578 on Weather, 0.3790/0.2578 on Traffic, and 0.1520/0.2465 on ECL. FARR achieves the lowest dataset-average MSE and MAE among the listed methods on all four datasets. At individual horizons, Time-MoE has lower ETTh1 error at $H=96$, and SE-LLM at $H=192$. Time-LLM has lower ECL MSE at $H=720$. FARR leads on dataset averages while individual-horizon rankings vary.

\subsection{Component ablations}
Table~\ref{tab:ablation} tests the input information and residual refinement mechanism. The local valid anchor is the uncorrected forecast from the evaluated checkpoint; SE-LLM in Table~\ref{tab:main} is a separately reported published baseline. Adjustment only applies Eq.~\eqref{eq:adjustment} without residual refinement. Matched MLP is a parameter-matched dual-endpoint MLP without explicit disagreement features, horizon-wise gating, or Stage 2. History only uses historical features and the horizon coordinate; single endpoint adds the valid forecast; disagreement only uses the disagreement features. Dual endpoint (Stage 1) retains all features without forecast adjustment.

\begin{table}[htbp]
\centering
\caption{Component ablations in standardized space, averaged over four horizons. Best values are bold.}
\label{tab:ablation}
\fontsize{10.1}{12.5}\selectfont
\setlength{\tabcolsep}{3.3pt}
\renewcommand{\arraystretch}{1.18}
\begin{tabular}{@{}l*{8}{r}@{}}
\toprule
&\multicolumn{2}{c}{ETTh1}&\multicolumn{2}{c}{Weather}&\multicolumn{2}{c}{Traffic}&\multicolumn{2}{c}{ECL}\\
\cmidrule(lr){2-3}\cmidrule(lr){4-5}\cmidrule(lr){6-7}\cmidrule(lr){8-9}
Variant&MSE&MAE&MSE&MAE&MSE&MAE&MSE&MAE\\
\midrule
Local valid anchor & 0.3900 & 0.4200 & 0.2300 & 0.2651 & 0.3865 & 0.2627 & 0.1605 & 0.2545 \\
Adjustment only & 0.3810 & 0.4151 & 0.2280 & 0.2661 & 0.3848 & 0.2609 & 0.1609 & 0.2540 \\
Matched MLP & 0.3796 & 0.4142 & 0.2258 & 0.2622 & 0.3835 & 0.2600 & 0.1584 & 0.2518 \\
History only & 0.3834 & 0.4170 & 0.2286 & 0.2668 & 0.3881 & 0.2635 & 0.1615 & 0.2549 \\
Single endpoint & 0.3818 & 0.4158 & 0.2270 & 0.2645 & 0.3856 & 0.2614 & 0.1602 & 0.2533 \\
Disagreement only & 0.3828 & 0.4164 & 0.2284 & 0.2659 & 0.3841 & 0.2607 & 0.1611 & 0.2545 \\
Dual endpoint (Stage 1) & \textbf{0.3770} & 0.4123 & 0.2230 & \textbf{0.2577} & 0.3821 & 0.2590 & 0.1562 & 0.2497 \\
\textbf{Full FARR} & 0.3773 & \textbf{0.4120} & \textbf{0.2208} & 0.2578 & \textbf{0.3790} & \textbf{0.2578} & \textbf{0.1520} & \textbf{0.2465} \\
\bottomrule
\end{tabular}
\end{table}

\begin{samepage}
Dual-endpoint refinement improves both metrics over the single-endpoint and disagreement-only variants on every dataset. On Weather, its MSE is 0.2230, compared with 0.2270 for single-endpoint refinement. Full FARR also outperforms the parameter-matched MLP in all eight dataset--metric comparisons. Against forecast adjustment alone, it reduces ECL MSE from 0.1609 to 0.1520. Stage 2 improves six of eight comparisons over Stage 1; ETTh1 MSE increases by 0.0003 and Weather MAE by 0.0001. The combined features improve refinement, while the additional gain from forecast adjustment varies across datasets.
\par\end{samepage}

\subsection{Implementation details}
Table~\ref{tab:settings} summarizes the settings. We use AdamW~\cite{adamw} with complete single-channel windows as batch samples. Stage 2 initializes from Stage 1 and jointly trains the refiner, horizon-wise gate, and forecast adjustment. Residual regularizers act before gating, whereas weight decay acts on parameters. Both endpoint passes reset seed 2021 and preserve sample order under deterministic inference. The final Stage-2 candidate is selected by joint validation improvement (Section~\ref{sec:selection}) before test evaluation.

\begin{table}[htbp]
\centering
\caption{Model hyperparameters and training settings.}
\label{tab:settings}
\fontsize{10.1}{12.5}\selectfont
\renewcommand{\arraystretch}{1.15}
\begin{tabular}{@{}ll@{}}
\toprule
Setting & Value\\
\midrule
Input / label / token length & 672 / 576 / 96\\
Prediction horizon $H$ & 96, 192, 336, 720\\
History / recurrence window & 168 / 24\\
Refiner architecture & $18\rightarrow128\rightarrow128\rightarrow1$\\
Residual output & $\tanh$; $[-1,1]$\\
Gate cap / initial logit & 0.20 / $-6$\\
SmoothL1 transition / $\alpha_r$ & 1.0 / 0.10\\
Smoothness / residual L2 weights & 0.01 / 0.01\\
Adjustment scale / offset bounds & 0.10 / 0.10\\
Batch size (train and evaluation) & 8192\\
Weight decay / gradient clipping & $10^{-4}$ / 1.0\\
Stage 1/2 learning rates & $10^{-3}$ / $10^{-4}$\\
Stage 1/2 epochs; patience & 10/5; 2/2\\
\bottomrule
\end{tabular}
\end{table}

\begin{samepage}
\subsection{Hyperparameter sensitivity}
We vary one factor at a time around $\lambda_{\max}=0.20$, $\alpha_r=0.10$, and $\beta_s=\beta_{\ell2}=0.01$. Alternatives are $\{0,0.05,0.10,0.40\}$ for $\lambda_{\max}$, $\{0,0.05,0.20\}$ for $\alpha_r$, and $\{0,0.001,0.05\}$ for each regularizer. Architecture, seed, optimizer, and training budget are fixed. Sensitivity runs use training and validation data only. Their checkpoint rule requires a strict MSE decrease and permits at most a 0.2\% MAE increase, unlike the joint strict improvement used for the main experiments.
\par\end{samepage}

\begin{figure}[htbp]
\centering
\includegraphics[width=0.76\textwidth]{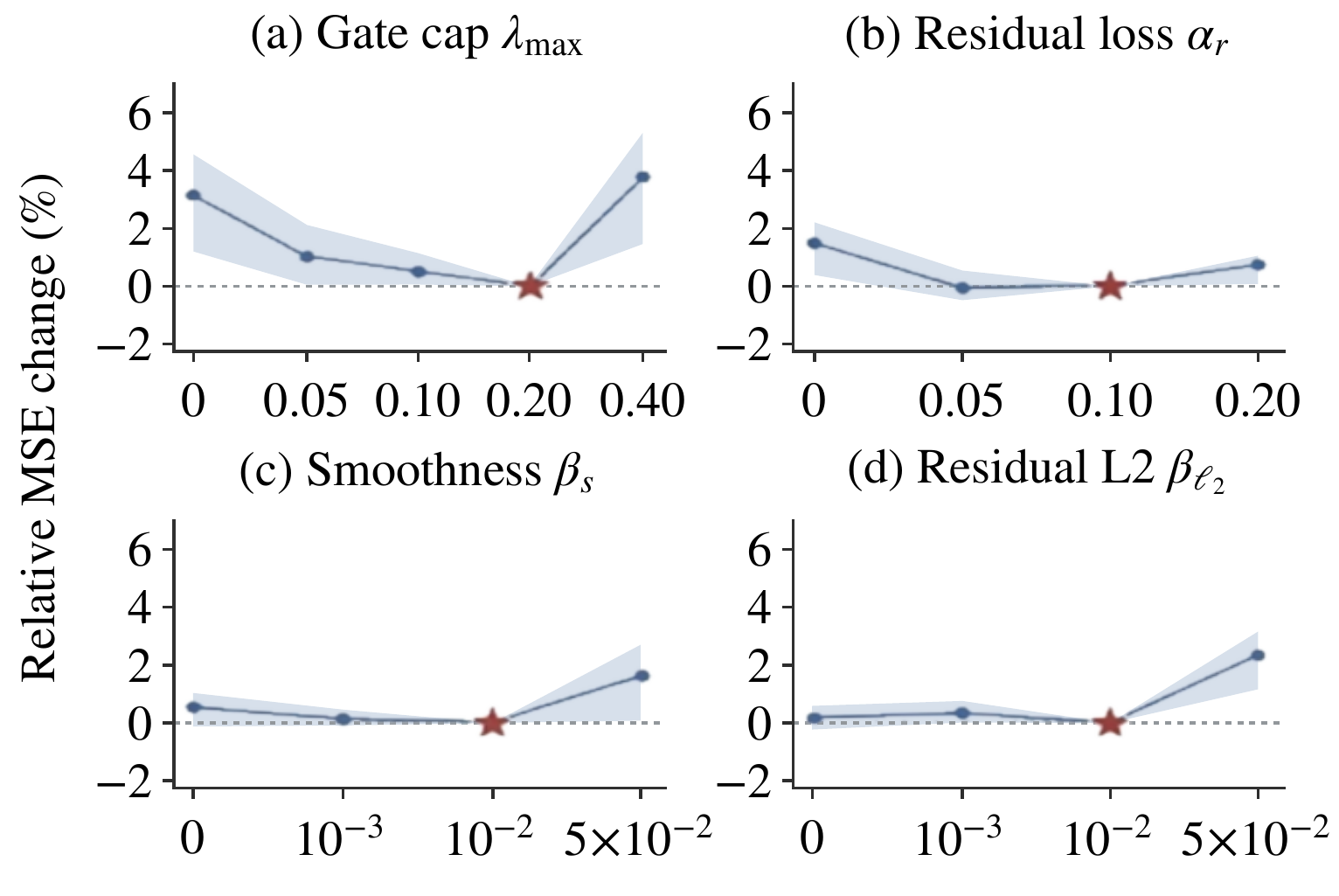}
\caption{Hyperparameter sensitivity. Lines and shading show mean relative MSE changes and the 25th--75th percentile range across included forecasting settings. Stars mark the normalization reference.}
\label{fig:sensitivity}
\end{figure}

In Figure~\ref{fig:sensitivity}, disabling residual refinement or increasing regularization produces larger MSE changes. Shading shows variation across the included forecasting settings. The zero-change reference is defined by normalization.

\subsection{Inference cost}
Table~\ref{tab:latency} reports the ETTh1 latency breakdown. Additional endpoint generation and subsequent correction add 21.01~ms and 1.08~ms, respectively, based on differences between execution-mode means. Full FARR takes 85.12~ms, 35.0\% more than the single-endpoint backbone. The standardized MSE reduction is $0.3900-0.3773=0.0127$, or 3.3\% relative to the local valid anchor, using the reported averages in Table~\ref{tab:ablation}.

\begin{table}[htbp]
\centering
\caption{ETTh1 mean inference latency (ms) and standardized MSE reduction relative to the local valid anchor. Latency increments are differences between execution-mode means.}
\label{tab:latency}
\fontsize{10.1}{12.5}\selectfont
\setlength{\tabcolsep}{8pt}
\renewcommand{\arraystretch}{1.2}
\begin{tabular}{@{}lrrrrr@{}}
\toprule
Dataset & \shortstack{Single\\backbone} & \shortstack{Extra\\endpoint} & \shortstack{Extra\\correction} & \shortstack{Full\\FARR} & \shortstack{MSE\\reduction}\\
\midrule
ETTh1 & 63.03 & 21.01 & 1.08 & 85.12 & 0.0127\\
\bottomrule
\end{tabular}
\end{table}

The additional endpoint accounts for 95.1\% of the 22.09~ms latency increase; subsequent correction contributes 4.9\%. The 19.4K--21.2K trainable parameters describe the refinement module, while deployment still requires two backbone passes.

\section{Conclusion}
FARR addresses the variable effects of auxiliary conditioning by learning residual corrections from matched valid and null forecasts and historical context. A lightweight refiner and bounded horizon-wise gate improve predictions while keeping the backbone frozen. Across four benchmarks, FARR achieves the lowest dataset-average MSE and MAE among the compared methods. Generating the additional endpoint incurs inference overhead. Future work will explore selective endpoint generation and distillation to retain the benefits of endpoint evidence with lower inference cost.

\section{Compliance with ethical standards}
This is a numerical simulation study for which no ethical approval was required.

\paragraph{Acknowledgments.} The authors have no relevant financial or nonfinancial interests to disclose.

\end{document}